\documentclass[runningheads]{llncs}
\usepackage[T1]{fontenc}
\usepackage{graphicx,verbatim}
\usepackage{xcolor} 
\usepackage{amsmath}
\usepackage{amssymb}

\makeatletter
\newcommand{\equalcontribmark}{%
  \textsuperscript{\@fnsymbol{1}}%
}
\makeatother

\begin{document}
\title{Synthetic training for long-tail haemorrhagic lesion segmentation in data-scarce settings}
%

\author{
Yuan Cao\inst{1}\thanks{Yuan Cao and Sumeet Dash contributed equally to this work.}
\orcidID{0009-0007-4323-0737} \and
Sumeet Dash\inst{1}\equalcontribmark \and
Antonia Zachariadis\inst{2} \and
Stefanie Schreiber\inst{2,3}\orcidID{0000-0003-4439-4374} \and
Katja Neumann\inst{2} \and
Jose Bernal\inst{1,4}\orcidID{0000-0003-3167-5134}
}

\authorrunning{Y. Cao et al.}
%
\institute{Department Artificial Intelligence in Biomedical Engineering (AIBE), Faculty of Engineering, Friedrich-Alexander-Universität Erlangen-Nürnberg (FAU), Nürnberger Strasse 74, 91052 Erlangen, Germany\\
\email{\{yuan.cao, sumeet.dash, jose.bernal\}@fau.de} \and
Otto von Guericke University Magdeburg, Medical Faculty, Department of Neurology, Leipziger Straße 44, 39120 Magdeburg, Germany\\
\email{antonia.zachariadis@st.ovgu.de}\\ \email{\{stefanie.schreiber, katja.neumann\}@med.ovgu.de}
\and
German Centre for Neurodegenerative Diseases (DZNE), Leipziger Straße 44, 39120 Magdeburg, Germany
\and
Institute for Neuroscience and Cardiovascular Research, Row Fogo Centre for Research into Ageing and The Brain, Department of Neuroimaging Sciences, The University of Edinburgh, 49 Little France Crescent, Edinburgh EH16 4TJ, UK}
  
\maketitle              
\begin{abstract}
Cerebral microbleeds (CMBs) and cortical superficial siderosis (cSS) are imaging markers of cerebral small vessel disease, but their automated segmentation is limited by the scarcity of positive cases and voxel-level annotations. We propose a synthetic training framework for long-tail haemorrhagic lesion segmentation that requires no real lesion annotations for training and leverages radiological description of the lesions. Starting from anatomical brain parcellations, the framework applies spatial augmentation and voxel resampling, procedurally inserts cSS and CMB labels using clinical priors on lesion location and morphology, and synthesises images through randomised intensity assignment, blurring, and Rician noise simulation. Models were trained on dynamically generated image--label pairs and evaluated against manual delineations in 10 cSS cases and 13 CMB cases. The proposed configurations outperformed classical filter baselines. For cSS, the hypointensity-constrained model achieved higher AUPRC and AUROC than the Frangi filter (AUPRC: 0.284 vs 0.083; AUROC: 0.907 vs 0.731). For CMBs, explicit synthesis of blood vessels as lesion mimics improved performance over the classical baseline (AUPRC: 0.538 vs 0.004; AUROC: 0.999 vs 0.968). These results support our proposal as a feasible strategy for data-scarce haemorrhagic lesion segmentation.

\keywords{Synthetic training \and Cortical superficial siderosis \and Cerebral microbleeds \and Segmentation \and Cerebral small vessel disease \and Brain MRI}

\end{abstract}
\section{Introduction}
Cerebral microbleeds (CMBs) and cortical superficial siderosis (cSS) are magnetic resonance imaging (MRI) markers of cerebral small vessel disease (CSVD), a major contributor to stroke, vascular cognitive impairment, dementia, and haemorrhagic adverse events during anti-amyloid therapies~\cite{duering2023neuroimaging,charidimou2022boston,sin2023anti,claus2025generalizability,lohner2025machine}. Accurate quantification of CMB and cSS burden is therefore important for characterising CSVD severity, stratifying haemorrhagic and dementia risk, assessing treatment eligibility, and monitoring disease progression and treatment safety.

When large, heterogeneous annotated datasets are available, standard supervised segmentation pipelines such as nnU-Net can achieve strong lesion-segmentation performance~\cite{isensee2021nnu}. This premise, however, breaks down for long-tail haemorrhagic lesion-segmentation tasks such as CMB and cSS segmentation. Positive cases are uncommon even in clinically relevant cohorts, with reported prevalences below 3\% for cSS~\cite{shams2016cortical} and below 20\% for CMBs in memory-clinic settings~\cite{cordonnier2006prevalence}. Moreover, lesion voxels are extremely sparse, and lesion burden can range from a few isolated voxels to disseminated involvement across the whole brain~\cite{duering2023neuroimaging}. Building representative training datasets therefore requires targeted expert case identification and voxel-level annotation across diverse cohorts, both of which are laborious and resource-intensive. For CMBs and cSS, such large, diverse, publicly available datasets remain unavailable, motivating alternative training strategies for data-scarce, long-tail lesion segmentation.

Synthetic training addresses this bottleneck by generating image--label pairs rather than requiring manual annotation for every training example. This strategy has proven effective when the target structures are already available as label maps, such as anatomical regions, tissue classes, or existing lesion masks~\cite{hoffmann2024anatomy,hoopes2022synthstrip,laso2024quantifying,billot2023synthseg,billot2023robust,Chalcroft2025}. For CMBs and cSS, however, voxel-level lesion labels are precisely missing, limiting the direct applicability of standard synthetic training frameworks.

In this paper, we propose to extend synthetic training to data-scarce haemorrhagic lesion segmentation by procedurally generating the missing CMB and cSS labels themselves using clinical and imaging priors on lesion morphology, anatomical distribution, MRI appearance, and acquisition variability. Evaluation against manual delineations across two datasets supports this approach for long-tail haemorrhagic lesion segmentation in data-scarce settings.

\section{Methods}
\begin{figure}
    \centering
    \includegraphics[width=0.99\linewidth]{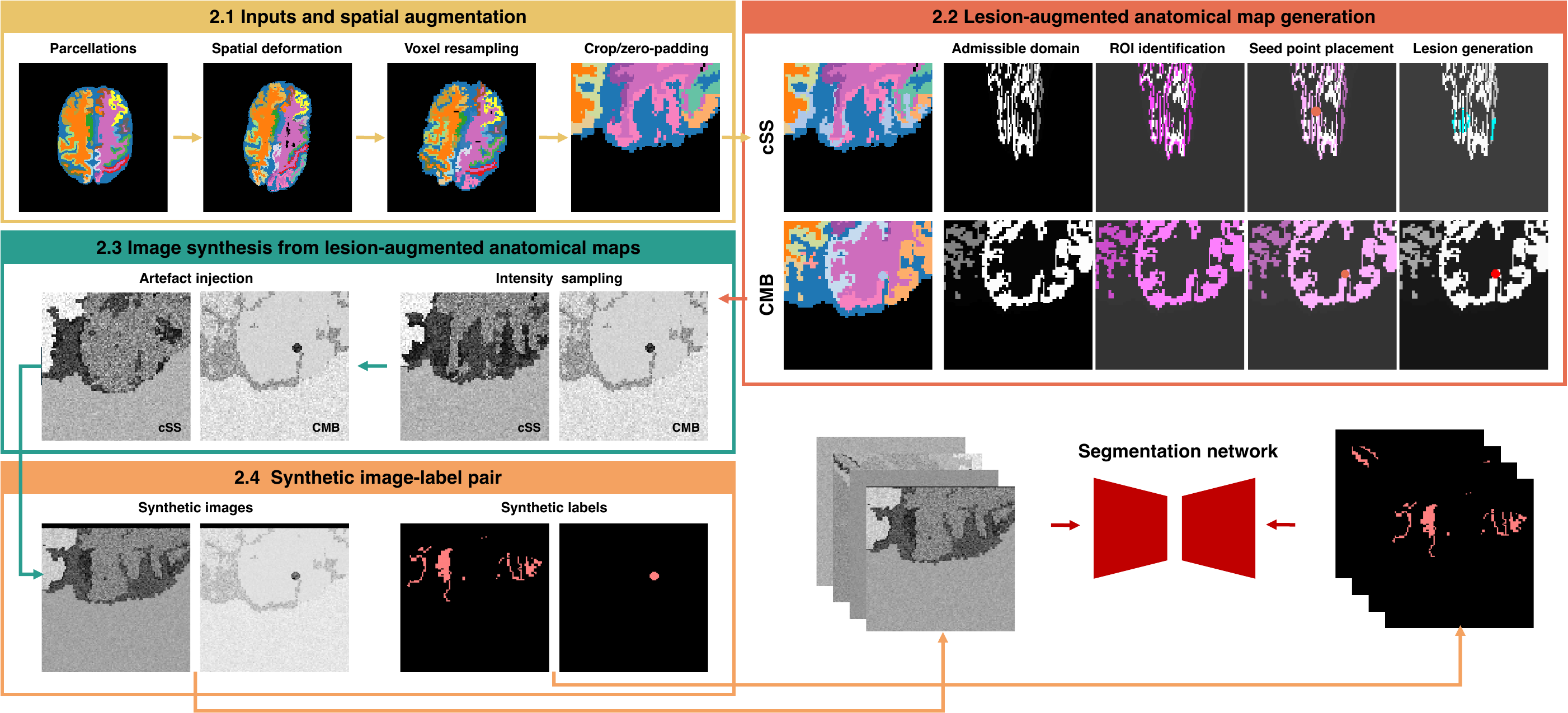}
    \caption{Proposed synthetic image generation pipeline for long-tail haemorrhagic lesion segmentation.\label{fig:pipeline}}
\end{figure}

The following sections describe the proposed synthetic data generation framework (Fig.~\ref{fig:pipeline}). In brief, anatomical parcellations are first spatially augmented to increase anatomical and voxel-size variability. cSS and CMB labels are then inserted using lesion-specific anatomical and morphological priors. The resulting lesion-augmented maps are converted into synthetic images through randomised intensity assignment, blurring, and noise simulation, and finally paired with binary lesion masks for network training.

\subsection{Inputs and spatial augmentation}

The data generation workflow starts from a pool of brain parcellations
$\{S_n\}_{n=1}^N$, where each parcellation is a label map
$S_n : \Omega_n \rightarrow \{1,\dots,K\}$. Here,
$\Omega_n \subset \mathbb{Z}^3$ denotes the discrete voxel lattice of subject $n$,
and $K$ is the number of anatomical classes before adding cSS/CMB labels. Although these maps contain no haemorrhagic lesion annotations, they provide the anatomical context required to place cSS and CMBs in clinically plausible locations.

At each training iteration, a parcellation $S_i$ is sampled at random and spatially augmented to increase anatomical variability and simulate native-space acquisition differences. The augmentation comprises a randomly parameterised affine transformation
$\phi_{\mathrm{aff}}(\cdot;\theta_{\mathrm{aff}})$, a nonlinear elastic deformation
$\phi_{\mathrm{nonlin}}(\cdot;\theta_{\mathrm{nonlin}})$, and voxel-size resampling
$\mathcal{R}_{\mathbf{r}_{\mathrm{res}}}$, where $\theta_{\mathrm{aff}}$ are the affine parameters, $\theta_{\mathrm{nonlin}}$ the deformation parameters, and $\mathbf{r}_{\mathrm{res}}$ the target voxel size.

Because transformation and resampling may change the lattice dimensions, a cropping or zero-padding operator $\mathcal{M}_{\Omega}$ is applied to enforce a consistent output lattice $\Omega \subset \mathbb{Z}^3$. The augmented parcellation is therefore defined as
\begin{equation}
S_i^{\mathrm{aug}} =
\mathcal{M}_{\Omega}
\mathcal{R}_{\mathbf{r}_{\mathrm{res}}}
\left[
S_i \circ
\left(
\phi_{\mathrm{aff}}(\cdot;\theta_{\mathrm{aff}})
\circ
\phi_{\mathrm{nonlin}}(\cdot;\theta_{\mathrm{nonlin}})
\right)
\right],
~
S_i^{\mathrm{aug}} : \Omega \rightarrow \{1,\dots,K\}.
\end{equation}

\subsection{Lesion-augmented anatomical map generation}

The STRIVE criteria~\cite{duering2023neuroimaging} define CMBs as small, quasi-spherical blood-product deposits within the brain parenchyma, whereas cSS reflects superficial blood-product deposition along the cortical surface and adjacent subarachnoid space. Lesion labels are therefore generated by combining lesion-specific priors on morphology and anatomical location with the augmented anatomical map $S_i^{\mathrm{aug}}$.

Let $\mathcal{L}_{\mathrm{GM}}\subset\{1,\dots,K\}$ and
$\mathcal{L}_{\mathrm{WM}}\subset\{1,\dots,K\}$ denote the grey matter and white matter label sets, respectively, and let $l_{\mathrm{CSF}}\in\{1,\dots,K\}$ denote the CSF label. For each lesion type $t\in\{\mathrm{cSS},\mathrm{CMB}\}$, an admissible domain $\mathcal{D}_i^t$ is first defined, seed points are sampled within this domain, and the resulting lesion mask $\Omega_i^t$ is inserted into the anatomical label map.

For cSS, the admissible domain is the CSF--cortical grey matter interface:
\begin{equation}
\mathcal{D}_i^{\mathrm{cSS}} =
\Gamma_i^{\mathrm{sur}} =
\left(\{x \in \Omega \mid S_i^{\mathrm{aug}}(x)=l_{\mathrm{CSF}}\} \oplus B\right)
\cap
\{x \in \Omega \mid S_i^{\mathrm{aug}}(x)\in \mathcal{L}_{\mathrm{GM}}\},
\end{equation}
where $\oplus$ denotes morphological dilation with structuring element $B$. For CMBs, the admissible domain is restricted to parenchymal tissue:
\begin{equation}
\mathcal{D}_i^{\mathrm{CMB}} =
\Omega_i^{\mathrm{par}} =
\{x \in \Omega \mid S_i^{\mathrm{aug}}(x)\in\mathcal{L}_{\mathrm{par}}\},
\qquad
\mathcal{L}_{\mathrm{par}}=\mathcal{L}_{\mathrm{GM}}\cup\mathcal{L}_{\mathrm{WM}}.
\end{equation}

Seed points $\mathcal{S}_i^t=\{s_1^t,\dots,s_{M^t}^t\}\subset\mathcal{D}_i^t$ are sampled with
$M^t\sim\mathcal{U}(M^t_{\min},M^t_{\max})$, and each seed is assigned a radius
$R_j^t\sim\mathcal{U}(R^t_{\min},R^t_{\max})$. Lesion geometry is then generated according to the lesion type:
\begin{equation}
\Omega_i^{\mathrm{cSS}} =
\bigcup_{j=1}^{M^{\mathrm{cSS}}}
\{x \in \Gamma_i^{\mathrm{sur}} \mid d_{\Gamma,\boldsymbol{\rho}}(x,s_j^{\mathrm{cSS}}) \leq R_j^{\mathrm{cSS}}\},
\end{equation}
for surface-constrained cSS, where $d_{\Gamma,\boldsymbol{\rho}}$ denotes distance along the cortical surface, and
\begin{equation}
\Omega_i^{\mathrm{CMB}} =
\bigcup_{j=1}^{M^{\mathrm{CMB}}}
\left\{
x \in \Omega_i^{\mathrm{par}}
\mid
\left\|\boldsymbol{\rho}\odot(x-s_j^{\mathrm{CMB}})\right\|_2 \leq R_j^{\mathrm{CMB}}
\right\},
\end{equation}
for approximately spherical CMBs in physical space, where $\boldsymbol{\rho}$ denotes voxel size.

Finally, the simulated lesion label is inserted into the augmented parcellation:
\begin{equation}
S_{i,t}^{\mathrm{final}}(x) =
\begin{cases}
l_t, & \text{if } x \in \Omega_i^t,\\
S_i^{\mathrm{aug}}(x), & \text{otherwise},
\end{cases}
\qquad
S_{i,t}^{\mathrm{final}} : \Omega \rightarrow \{1,\dots,K\}\cup\{l_t\}.
\end{equation}

\subsection{Image synthesis from lesion-augmented anatomical maps}
The intensity model assigns each label $l$ an intensity parameter $\mu_l$. For anatomical labels
$k\in\mathcal{L}_{\mathrm{anat}}=\{1,\dots,K\}$, intensities are sampled as
$\mu_k\sim\mathcal{U}(0,1)$. For haemorrhagic lesion labels, intensities are constrained by an imaging prior. On T2*-weighted imaging and SWI, blood-breakdown products cause local signal loss, making CMBs and cSS hypointense relative to neighbouring tissues~\cite{charidimou2022boston,duering2023neuroimaging}. Therefore, for lesion labels
$h\in\mathcal{L}_{\mathrm{hem}}=\{l_{\mathrm{CMB}},l_{\mathrm{cSS}}\}$, intensities are sampled under the constraint
\begin{equation}
\max_{h\in\mathcal{L}_{\mathrm{hem}}}\mu_h
<
\min_{k\in\mathcal{L}_{\mathrm{anat}}}\mu_k .
\end{equation}
The initial synthetic image is then obtained by replacing each voxel label in the lesion-augmented map with its sampled intensity $I_0(x)=\mu_{S_i^{\mathrm{final}}(x)},~ x\in\Omega$.

To mimic partial-volume effects and acquisition noise, the initial synthetic image $I_0$ is first smoothed with a Gaussian kernel $G_{\sigma_{\mathrm{blur}}}$ and then corrupted with Rician noise. Noise is simulated by adding independent complex Gaussian components to the $k$-space representation, with variance determined by a randomly sampled
$\mathrm{SNR}_{\mathrm{dB}}\sim\mathcal{U}(\mathrm{SNR}_{\min},\mathrm{SNR}_{\max})$. The resulting image is obtained by inverse Fourier transformation and magnitude reconstruction:
\begin{equation}
I_{\mathrm{final}}(x)
=
\left|
\mathcal{F}^{-1}
\left(
\mathcal{F}(I_0 * G_{\sigma_{\mathrm{blur}}})
+
\eta_{\mathrm{real}}
+
i\eta_{\mathrm{imag}}
\right)(x)
\right|,
\end{equation}
where $\eta_{\mathrm{real}},\eta_{\mathrm{imag}}\sim\mathcal{N}(0,\sigma_{\mathrm{noise}}^2)$ are independent Gaussian noise components, with $
\sigma_{\mathrm{noise}}
=
\mu_{\mathrm{signal}}/10^{\mathrm{SNR}_{\mathrm{dB}}/20}$.

\subsection{Final synthetic image-label pair}
Training synthetic image--mask pairs $(I_{\mathrm{final}},Y)$ are constructed by using
$I_{\mathrm{final}}$ as the input image and binarising $S_i^{\mathrm{final}}$ to isolate the target cSS/CMB label.

\section{Training and evaluation}
\subsection{Training}
\textbf{Brain parcellations.} We used a pool of 840 brain parcellations derived from Alzheimer's Disease Neuroimaging Initiative (ADNI) scans. Parcellations were generated with SynthSeg~\cite{billot2023synthseg} and SIAM~\cite{valabregue2026siam}, which provides labels to guide cSS and CMB placement, respectively.

\textbf{Generation parameters.} 
Brain parcellations were spatially augmented with affine transformations (scaling $\pm20\%$, rotation $\pm15^\circ$, shearing $\pm0.012$) and nonlinear elastic deformations (displacement-field std 3.0, control-point scale 0.0625). To mimic acquisition variability, volumes were randomly resampled to isotropic resolutions of 0.5--1.2 mm or anisotropic resolutions with 0.5--1.2 mm in-plane and 1.2--4.0 mm through-plane spacing. For cSS, 3--20 lesions were generated from sampled seeds using geodesic propagation, with a base cost of 1.0, noise weight of 0.25, and $\boldsymbol{\rho}\sim\mathcal{U}(10, 30)$ voxels. For CMB, radii were sampled uniformly from 1--5\,mm, with 1--4 lesions per crop. In both cases, $SNR_{dB}$ was drawn from a uniform distribution $\mathcal{U}(10~dB, 40~dB)$.

\textbf{Segmentation network.} We used a five-level 3D U-Net backbone. Each block comprised two convolutional layers with batch normalisation and ELU activations. The encoder used max-pooling for downsampling, the decoder used upsampling, and skip connections preserved local spatial detail. Voxel-wise predictions were produced with a final softmax layer. The model was implemented in PyTorch/MONAI and trained for 70 epochs, with 1,000 on-the-fly synthesised image--label pairs per epoch and a batch size of $2$. Optimisation used Adam with a learning rate of $10^{-5}$. To address severe class imbalance, training minimised a soft Dice loss with a squared denominator and dynamic inverse-volume weighting~\cite{billot2023synthseg,bitar2025drips}. Training took approximately 24 hours on an NVIDIA A100 GPU.

\subsection{Evaluation}
\textbf{Test cases and ground truth.}
For cSS, evaluation used an in-house test set of 10 patients with confirmed cSS (6 females; median age 74.5 [IQR 67--78] years) scanned with susceptibility- or T2*-weighted MRI. Lesions were manually delineated by a trained investigator and validated by an experienced neuroradiologist. For CMBs, evaluation used 13 T2*-weighted scans from the ALFA cohort made available through the MICCAI 2021 VALDO Challenge~\cite{sudre2024valdo}, under the corresponding ethics approval and data use agreements.

\textbf{Ablation experiments.}
To quantify the effect of key generation choices, we performed ablation experiments. First, we compared fully randomised label intensities with the proposed intensity prior that enforces cSS and CMB hypointensity, testing whether preserving haemorrhagic signal dropout improves segmentation beyond anatomical and morphological randomisation alone. Second, we compared image-only input with input augmented by a Laplacian filter response, testing whether an explicit cue for dark lesion-like structures improves learning. Under the adopted sign convention, hypointense structures yield positive Laplacian responses, highlighting candidate CMB and cSS patterns while preserving the original synthetic image information. Third, we also evaluated lesion-specific refinements motivated by observed failure modes. For CMBs, false positives frequently occurred along blood vessels; we therefore tested whether including blood vessel labels from SIAM during synthesis improved robustness to hypointense vessel-like structures. For cSS, errors tended to occur at the CSF--cortical boundary. We therefore tested whether a two-class target formulation improved separation between cSS and CSF.

\textbf{Comparison against other methods.} For cSS, no established method provides fully automated segmentation. The closest existing approach is a semi-automated Frangi-filter-based method~\cite{van2023quantitative}, which we therefore used as comparator. For CMBs, we also used the Frangi as the corresponding classical baseline.

\textbf{Evaluation metrics.} Segmentation performance was evaluated using threshold-free metrics: area under the precision--recall curve (AUPRC) and area under the receiver operating characteristic curve (AUROC). These metrics allow  comparison with classical non-probabilistic baselines, such as the Frangi filter.

\section{Results and discussion}

\begin{figure}
    \centering
    \includegraphics[width=0.8\linewidth,trim={0.2cm 0.5cm 0.3cm 0.2cm},
  clip]{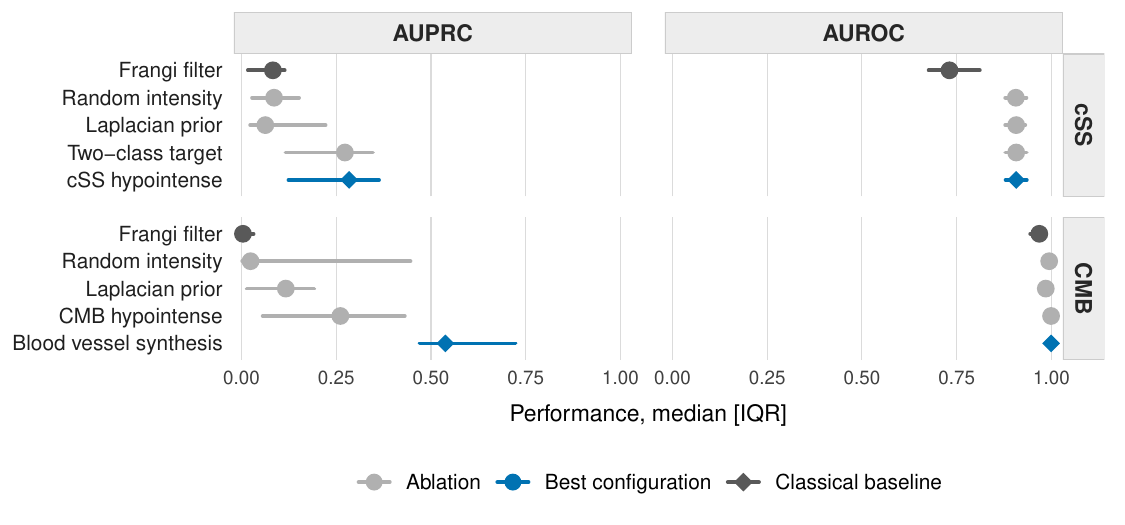}
    \caption{Quantitative evaluation of framework components for cSS and CMB segmentation. Forest plots show median AUPRC and AUROC with interquartile ranges for each approach. The proposed hypointensity prior improved performance for both lesion types, while explicit blood-vessel synthesis led to the best CMB performance.\label{fig:results}}
\end{figure}

\begin{figure}
\centering
\includegraphics[width=0.99\textwidth,trim={0.5cm 2cm 0.5cm 2cm},
  clip]{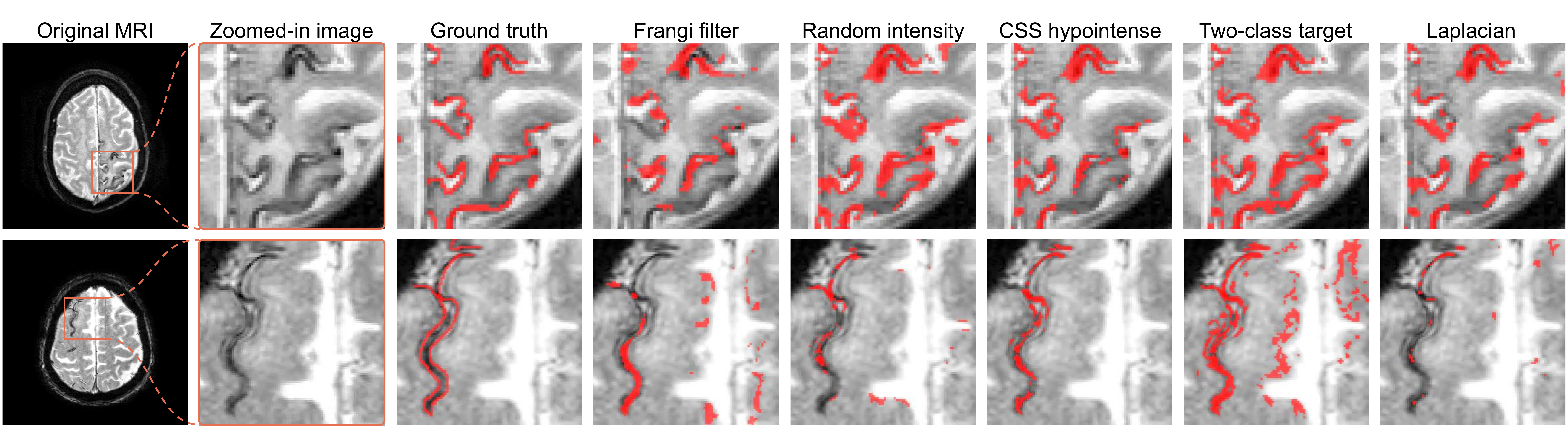}
\includegraphics[width=0.99\textwidth,trim={0.1cm 0.1cm 0.1cm 0.1cm},
  clip]{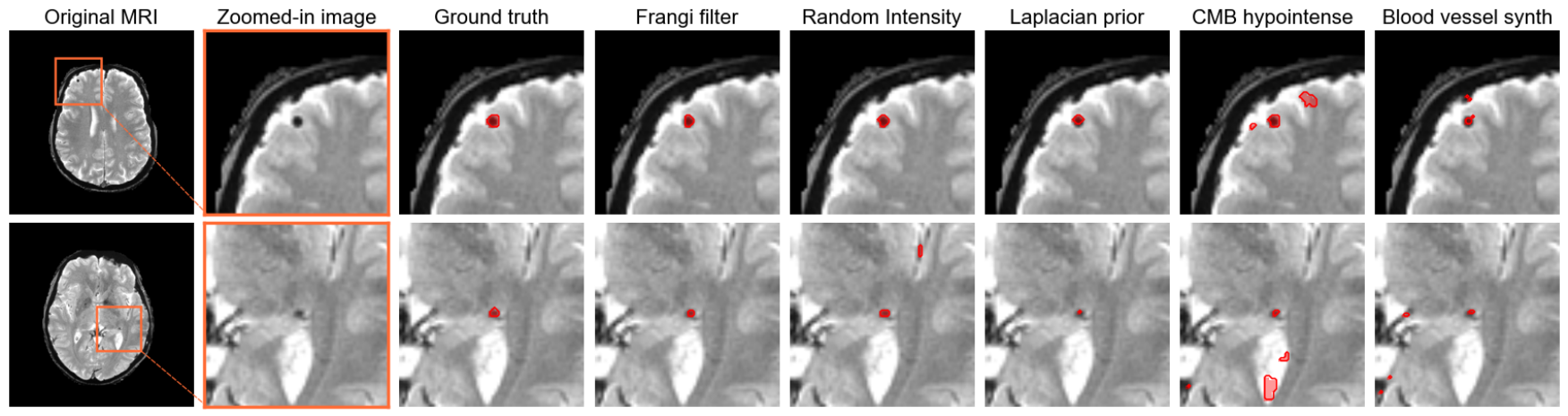}
\caption{Visual comparison of segmentation performance. Columns show, from left to right, the original MRI scan, zoomed-in view, manual ground-truth annotation, Frangi filter output, and outputs from the four ablation configurations: random intensity sampling, Laplacian prior, cSS/CMB hypointensity constraint, and lesion-specific refinement. The lesion-specific refinement corresponds to a two-class target for cSS and blood-vessel synthesis for CMBs.}\label{fig:illustration}
\end{figure}

Results are shown in Figs.~\ref{fig:results}--\ref{fig:illustration} and discussed in the following sections.

\textbf{Random intensities vs intensity prior.} Fully randomised intensity sampling yielded substantially lower performance than the proposed hypointensity prior for both cSS and CMB segmentation, especially for AUPRC (cSS: 0.086 [IQR 0.028--0.152] vs 0.284 [IQR 0.124--0.363]; CMB: 0.024 [IQR 0.002--0.446] vs 0.261 [IQR 0.056--0.431]), while AUROC remained largely unchanged. This suggests that enforcing haemorrhagic-lesion hypointensity improves precision in the sparse lesion class by ranking true lesion voxels higher among lesion-like candidates and reducing false positives. In contrast, fully randomised intensities make the synthetic task less aligned with blood-sensitive MRI.

\textbf{Image prior only vs image and Laplacian prior.} Although the Laplacian channel was intended to provide an explicit cue for local hypointense structures, it also enhanced non-lesion features, including vessel boundaries and sulcal interfaces. Consequently, adding the Laplacian prior degraded performance relative to the image prior alone for both lesion types (cSS AUPRC: 0.063 [IQR 0.023--0.222] vs 0.284 [IQR 0.124--0.363]; CMB AUPRC: 0.117 [IQR 0.014--0.192] vs 0.261 [IQR 0.056--0.431]). These findings suggest that the constrained synthetic images already encoded sufficient intensity and morphological information, whereas the Laplacian channel introduced a less specific representation that impaired transfer to real data.

\textbf{cSS-specific refinements.}
For cSS, visual inspection showed that errors frequently occurred at the CSF--cortical boundary, particularly when cSS was adjacent to sulcal CSF. To address this failure mode, we tested an alternative target formulation in which CSF was modelled explicitly as a separate class, alongside background and cSS, rather than training only a binary cSS-versus-background model. However, this modification produced little numerical change. AUPRC was similar for the binary cSS target and the explicit-CSF formulation (0.284 [IQR 0.124--0.363] vs 0.273 [IQR 0.116--0.347]). This suggests that most boundary errors were not primarily caused by the absence of an explicit CSF class, but may instead reflect the intrinsic ambiguity of thin hypointense structures at the cortical surface, partial-volume effects, and the limited spatial separation between cSS and adjacent CSF.

\textbf{CMB-specific refinements.} For CMBs, visual inspection showed that false positives often occurred along cerebral blood vessels, which can also appear hypointense on blood-sensitive MRI. To address this failure mode, we tested whether explicit vessel modelling during synthesis improved robustness. Because SynthSeg does not provide vascular labels, SIAM parcellations were used to introduce vessels as a separate anatomical class in the synthetic label maps. This CMB-specific refinement substantially improved performance, increasing AUPRC from 0.261 [IQR 0.056--0.431] to 0.538 [IQR 0.470--0.723]. This suggests that explicitly modelling relevant lesion mimics during synthetic training is important for reducing false positives.

\textbf{Comparison against other methods} Compared with the classical filter baseline, the best synthetic-training configurations showed higher median AUPRC and AUROC for both lesion types. For cSS, the hypointensity-constrained model achieved higher AUPRC than the Frangi filter (0.284 [IQR 0.124--0.363] vs 0.083 [IQR 0.017--0.114]) and higher AUROC (0.907 [IQR 0.879--0.935] vs 0.731 [IQR 0.676--0.811]). For CMBs, the model trained with blood-vessel synthesis achieved higher AUPRC than the classical filter baseline (0.538 [IQR 0.470--0.723] vs 0.004 [IQR 0.002--0.032]) and higher AUROC (0.999 [IQR 0.997--1.000] vs 0.968 [IQR 0.944--0.978]). Together, these results indicate better discrimination of the sparse positive class, with a more favourable balance between true-positive lesion voxels and false-positive detections, as reflected by AUPRC. The higher AUROC further suggests improved lesion-versus-background ranking across thresholds, although this metric is less sensitive to false positives in highly imbalanced voxel-wise segmentation.

\section{Conclusion}
We introduced a synthetic training strategy for long-tail haemorrhagic lesion segmentation and demonstrated its feasibility for CMB and cSS quantification. By procedurally generating lesion labels from anatomical and imaging priors, the proposed framework enabled model training without real voxel-level lesion annotations. Experimental results showed agreement with manual delineations across two datasets. Future work will evaluate the approach in larger, more heterogeneous cohorts and extend it from segmentation to lesion detection.

    

\begin{credits}
\subsubsection{\ackname} None.

\subsubsection{\discintname} The authors have no competing interests to declare that are relevant to the content of this article. 
\end{credits}

%
%
%
\bibliographystyle{splncs04}
\bibliography{mybibliography}

@article{duering2023neuroimaging,
  title={Neuroimaging standards for research into small vessel disease—advances since 2013},
  author={Duering, Marco and Biessels, Geert Jan and Brodtmann, Amy and Chen, Christopher and Cordonnier, Charlotte and de Leeuw, Frank-Erik and Debette, St{\'e}phanie and Frayne, Richard and Jouvent, Eric and Rost, Natalia S and others},
  journal={The Lancet Neurology},
  volume={22},
  number={7},
  pages={602--618},
  year={2023},
  publisher={Elsevier}
}

@article{claus2025generalizability,
  title={Generalizability of trial criteria on amyloid-lowering therapy against Alzheimer’s disease to individuals with mild cognitive impairment or early Alzheimer’s disease in the general population},
  author={Claus, Jacqueline J and Vom Hofe, Ilse and van Ijlzinga Veenstra, Annekee and Licher, Silvan and Seelaar, Harro and de Jong, Frank J and Neitzel, Julia and Vernooij, Meike W and Ikram, M Arfan and Wolters, Frank J},
  journal={European Journal of Epidemiology},
  volume={40},
  number={3},
  pages={327--337},
  year={2025},
  publisher={Springer}
}

@article{sin2023anti,
  title={Anti-amyloid therapy, AD, and ARIA: untangling the role of CAA},
  author={Sin, Mo-Kyung and Zamrini, Edward and Ahmed, Ali and Nho, Kwangsik and Hajjar, Ihab},
  journal={Journal of clinical medicine},
  volume={12},
  number={21},
  pages={6792},
  year={2023},
  publisher={MDPI}
}

@article{charidimou2022boston,
  title={The Boston criteria version 2.0 for cerebral amyloid angiopathy: a multicentre, retrospective, MRI--neuropathology diagnostic accuracy study},
  author={Charidimou, Andreas and Boulouis, Gregoire and Frosch, Matthew P and Baron, Jean-Claude and Pasi, Marco and Albucher, Jean Francois and Banerjee, Gargi and Barbato, Carmen and Bonneville, Fabrice and Brandner, Sebastian and others},
  journal={The Lancet Neurology},
  volume={21},
  number={8},
  pages={714--725},
  year={2022},
  publisher={Elsevier}
}

@article{isensee2021nnu,
  title={nnU-Net: a self-configuring method for deep learning-based biomedical image segmentation},
  author={Isensee, Fabian and Jaeger, Paul F and Kohl, Simon AA and Petersen, Jens and Maier-Hein, Klaus H},
  journal={Nature methods},
  volume={18},
  number={2},
  pages={203--211},
  year={2021},
  publisher={Nature Publishing Group US New York}
}

@article{hoffmann2024anatomy,
  title={Anatomy-aware and acquisition-agnostic joint registration with SynthMorph},
  author={Hoffmann, Malte and Hoopes, Andrew and Greve, Douglas N and Fischl, Bruce and Dalca, Adrian V},
  journal={Imaging Neuroscience},
  volume={2},
  pages={imag--2},
  year={2024},
  publisher={MIT Press 255 Main Street, 9th Floor, Cambridge, Massachusetts 02142, USA~…}
}

@article{hoopes2022synthstrip,
  title={SynthStrip: skull-stripping for any brain image},
  author={Hoopes, Andrew and Mora, Jocelyn S and Dalca, Adrian V and Fischl, Bruce and Hoffmann, Malte},
  journal={NeuroImage},
  volume={260},
  pages={119474},
  year={2022},
  publisher={Elsevier}
}

@inproceedings{laso2024quantifying,
  title={Quantifying white matter hyperintensity and brain volumes in heterogeneous clinical and low-field portable MRI},
  author={Laso, Pablo and Cerri, Stefano and Sorby-Adams, Annabel and Guo, Jennifer and Mateen, Farrah and Goebl, Philipp and Wu, Jiaming and Liu, Peirong and Li, Hongwei Bran and Young, Sean I and others},
  booktitle={2024 IEEE International Symposium on Biomedical Imaging (ISBI)},
  pages={1--5},
  year={2024},
  organization={IEEE}
}

@article{billot2023synthseg,
  title={SynthSeg: Segmentation of brain MRI scans of any contrast and resolution without retraining},
  author={Billot, Benjamin and Greve, Douglas N and Puonti, Oula and Thielscher, Axel and Van Leemput, Koen and Fischl, Bruce and Dalca, Adrian V and Iglesias, Juan Eugenio and others},
  journal={Medical image analysis},
  volume={86},
  pages={102789},
  year={2023},
  publisher={Elsevier}
}

@article{billot2023robust,
  title={Robust machine learning segmentation for large-scale analysis of heterogeneous clinical brain MRI datasets},
  author={Billot, Benjamin and Magdamo, Colin and Cheng, You and Arnold, Steven E and Das, Sudeshna and Iglesias, Juan Eugenio},
  journal={Proceedings of the National Academy of Sciences},
  volume={120},
  number={9},
  pages={e2216399120},
  year={2023},
  publisher={National Academy of Sciences}
}

@article{Chalcroft2025,
  title = {Synthetic Data for Robust Stroke Segmentation},
  volume = {3},
  ISSN = {2766-905X},
  url = {http://dx.doi.org/10.59275/j.melba.2025-f3g6},
  DOI = {10.59275/j.melba.2025-f3g6},
  number = {August 2025},
  journal = {Machine Learning for Biomedical Imaging},
  publisher = {Machine Learning for Biomedical Imaging},
  author = {Chalcroft, Liam and Pappas, Ioannis and Price, Cathy J. and Ashburner, John},
  year = {2025},
  month = aug,
  pages = {317–346}
}

@misc{valabregue2026siam,
  title        = {{SIAM}: Head and Brain {MRI} Segmentation from Few High-Quality Templates via Synthetic Training},
  author       = {Valabregue, Romain and Khemir, Ines and Badinet, Eric and Rousseau, Fran{\c{c}}ois and Auzias, Guillaume and Dorent, Reuben},
  howpublished = {arXiv preprint arXiv:2605.02737},
  year         = {2026}
}

@article{sudre2024valdo,
  title={Where is VALDO? VAscular Lesions Detection and segmentatiOn challenge at MICCAI 2021},
  author={Sudre, Carole H and Van Wijnen, Kimberlin and Dubost, Florian and Adams, Hieab and Atkinson, David and Barkhof, Frederik and Birhanu, Mahlet A and Bron, Esther E and Camarasa, Robin and Chaturvedi, Nish and others},
  journal={Medical Image Analysis},
  volume={91},
  pages={103029},
  year={2024},
  publisher={Elsevier}
}

@article{shams2016cortical,
  title={Cortical superficial siderosis: prevalence and biomarker profile in a memory clinic population},
  author={Shams, Sara and Martola, Juha and Charidimou, Andreas and Cavallin, Lena and Granberg, Tobias and Shams, Mana and Forslin, Yngve and Aspelin, Peter and Kristoffersen-Wiberg, Maria and Wahlund, Lars-Olof},
  journal={Neurology},
  volume={87},
  number={11},
  pages={1110--1117},
  year={2016},
  publisher={Lippincott Williams \& Wilkins Hagerstown, MD}
}

@article{cordonnier2006prevalence,
  title={Prevalence and severity of microbleeds in a memory clinic setting},
  author={Cordonnier, C and Van der Flier, WM and Sluimer, JD and Leys, D and Barkhof, F and Scheltens, PH},
  journal={Neurology},
  volume={66},
  number={9},
  pages={1356--1360},
  year={2006},
  publisher={Lippincott Williams \& Wilkins}
}

@article{van2023quantitative,
  title={Quantitative measurement of cortical superficial siderosis in cerebral amyloid angiopathy},
  author={van Harten, TW and Koemans, EA and Voigt, S and Rasing, I and van Osch, MJP and van Walderveen, MAA and Wermer, MJH},
  journal={NeuroImage: Clinical},
  volume={38},
  pages={103447},
  year={2023},
  publisher={Elsevier}
}

@misc{bitar2025drips,
  title  = {{DRIPS}: Domain Randomisation for Image-based Perivascular Spaces Segmentation},
  author = {Bitar, Luna and
            D{\'i}az, Mario and
            Duarte Coello, Roberto and
            Vald{\'e}s-Hern{\'a}ndez, Maria d.C. and
            Mattern, Hendrik and
            Neumann, Katja and
            Pfister, Malte and
            Beck, Carolin and
            Mai, Huy Trong and
            Fuchs, Erelle and
            Tang, Serena and
            Tosun, Duygu and
            Besteher, Bianca and
            Rockt{\"a}schel, Tonia and
            Reuken, Philipp A. and
            Stallmach, Andreas and
            Opel, Nils and
            Gaser, Christian and
            Walter, Martin and
            D{\"o}rner, Marc and
            Arndt, Philipp and
            Behme, Daniel and
            Piechowiak, Christiane and
            Lading, Yves and
            M{\"u}ller, Patrick and
            Braun-Dullaeus, R{\"u}diger and
            Meuth, Sven G. and
            {{Alzheimer's Disease Neuroimaging Initiative}} and
            Wardlaw, Joanna M. and
            Schreiber, Stefanie and
            Trujillo, Maria and
            D{\"u}zel, Emrah and
            Ziegler, Gabriel and
            Bernal, Jose},
  howpublished = {medRxiv preprint 2025.10.22.25337423},
  year         = {2025}
}

@article{lohner2025machine,
  title={Machine learning applications in vascular neuroimaging for the diagnosis and prognosis of cognitive impairment and dementia: a systematic review and meta-analysis},
  author={Lohner, Valerie and Badhwar, Amanpreet and Detcheverry, Flavie E and Garc{\'\i}a, Cindy L and Gellersen, Helena M and Khodakarami, Zahra and Lattmann, Ren{\'e} and Li, Rui and Low, Audrey and Mazo, Claudia and others},
  journal={Alzheimer's research \& therapy},
  volume={17},
  number={1},
  pages={183},
  year={2025},
  publisher={Springer}
}

\end{document}